\documentclass[conference]{IEEEtran}
\usepackage{subfigure}
\usepackage{listings}
\usepackage[table,xcdraw]{xcolor}
\usepackage{amsmath}
\usepackage{pifont}
\usepackage{graphicx}
\usepackage{amssymb}
\definecolor{mydarkblue}{rgb}{0,0.08,0.65}
\usepackage[colorlinks=true,
    linkcolor=mydarkblue,
    citecolor=mydarkblue,
    filecolor=mydarkblue,
    urlcolor=mydarkblue]{hyperref} 
\usepackage{cleveref}
\usepackage{soul}
\usepackage[shortlabels]{enumitem}
\usepackage{url}
\usepackage{listings}
\usepackage{color}
\usepackage{tikz}
\usepackage{balance}
\usepackage{multirow}
\usepackage{comment}
\usepackage{amsmath}
\usepackage{graphicx}
\usepackage{wrapfig,lipsum,booktabs}
\usepackage{titlesec}
\usepackage[frozencache,cachedir=.]{minted}
\usepackage[symbol]{footmisc}
\usepackage{tablefootnote}
\usepackage{tabularx}

\usepackage{graphicx}
\usepackage{subcaption}
\usepackage{multicol}

\usepackage{dblfloatfix}
\usepackage{natbib}

\definecolor{codegreen}{rgb}{0,0.6,0}
\definecolor{codegray}{rgb}{0.5,0.5,0.5}
\definecolor{codepurple}{rgb}{0.58,0,0.82}
\definecolor{backcolour}{rgb}{0.95,0.95,0.92}

\makeatletter
\def\blfootnote{\xdef\@thefnmark{}\@footnotetext}
\makeatother

\lstdefinestyle{mystyle}{
  backgroundcolor=\color{backcolour},   commentstyle=\color{codegreen},
  keywordstyle=\color{magenta},
  numberstyle=\tiny\color{codegray},
  stringstyle=\color{codepurple},
  basicstyle=\ttfamily\footnotesize,
  breakatwhitespace=false,         
  breaklines=true,                 
  captionpos=b,                    
  keepspaces=true,                 
  numbers=left,                    
  numbersep=5pt,                  
  showspaces=false,                
  showstringspaces=false,
  showtabs=false,                  
  tabsize=2,
}

\AtBeginDocument{%
  \providecommand\BibTeX{{%
    \normalfont B\kern-0.5em{\scshape i\kern-0.25em b}\kern-0.8em\TeX}}}

\IEEEoverridecommandlockouts
\begin{document}

\makeatletter
  \def\title@font{\Large}
  \let\ltx@maketitle\@maketitle
  \def\@maketitle{\bgroup%
    \let\ltx@title\@title%
    \def\@title{\resizebox{\textwidth}{!}{%
      \mbox{\title@font\ltx@title}%
    }}%
    \ltx@maketitle%
  \egroup}
\makeatother

\title{The Zamba2 Suite: Technical Report}
\author{Paolo Glorioso$ \quad$ Quentin Anthony$ \quad$ Yury Tokpanov$ \quad$   \\ $\quad$  Anna Golubeva$ \quad$ Vasudev Shyam$ \quad$ James Whittington$ \quad$  Jonathan Pilault$  \quad $ Beren Millidge \\
{
\small
\{paolo, quentin, yury, anna, vasu, james, jonathan, beren\}@zyphra.com
}\\
{}\\
{
 Zyphra
 \small
}\\
{
\small
 Palo Alto, CA
}
}
\maketitle

\setcounter{page}{1}

\begin{abstract}

In this technical report, we present the Zamba2 series -- a suite of 1.2B, 2.7B, and 7.4B parameter hybrid Mamba2-transformer models that achieve state of the art performance against the leading open-weights models of their class, while achieving substantial gains in inference latency, throughput, and memory efficiency. The Zamba2 series builds upon our initial work with Zamba1-7B, optimizing its architecture, training and annealing datasets, and training for up to three trillion tokens. We provide open-source weights for all models of the Zamba2 series as well as instruction-tuned variants that are strongly competitive against comparable instruct-tuned models of their class. We additionally open-source the pretraining dataset, which we call Zyda-2, used to train the Zamba2 series of models. The models and datasets used in this work are openly available at \url{https://huggingface.co/Zyphra}

\end{abstract}

\section{Introduction}

\begin{figure*}
    \centering
    \includegraphics[width=\linewidth]{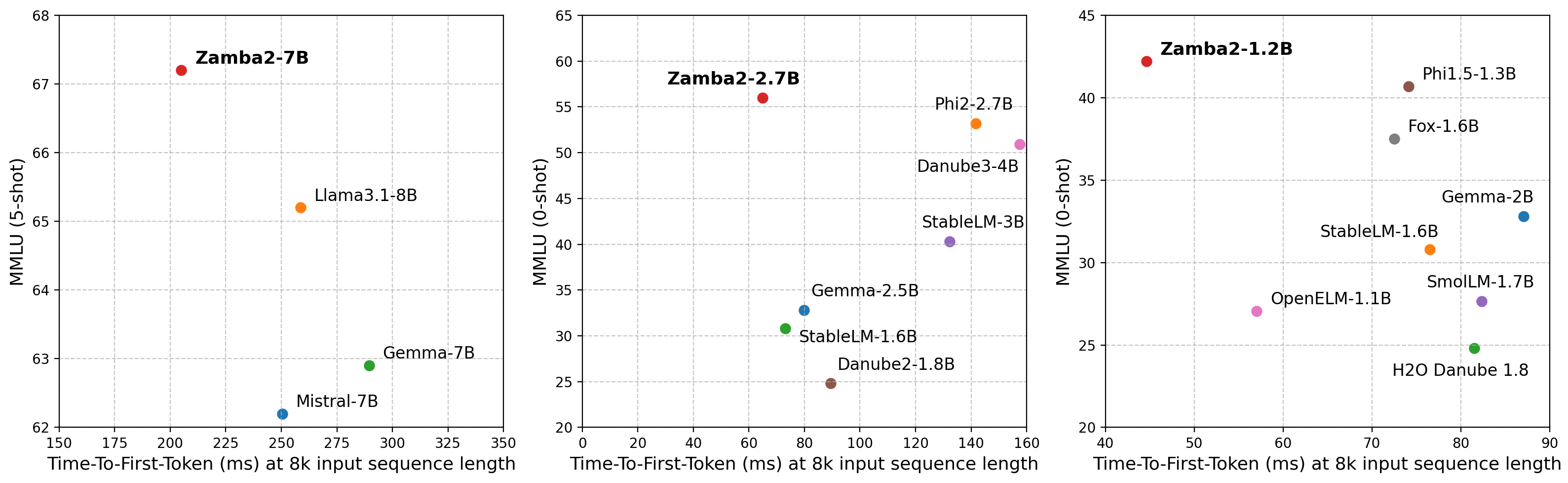}
    \caption{Performance (MMLU 5-shot or 0-shot) vs time-to-first-token for the Zamba2 series models vs leading competing models. Due to its novel Zamba2 architecture, our series of models significantly outperforms others in both quality and latency.}
    \label{fig:enter-label}
\end{figure*}

The transformer architecture \citep{vaswani2017attention} has revolutionized almost all fields of machine learning since its introduction seven years ago. The transformer is the first universally expressive sequence mixer made practical to scale by its efficient parallelization on GPUs. This efficiency has allowed transformer models to be scaled up by many orders of magnitude since their discovery, using unprecedented computational resources, leading to vastly enhanced performance and unlocking a wide range of capabilities. While transformers are an extremely powerful architecture at modeling, their expressivity comes at the cost of high compute and memory overheads. Specifically, a critical limitation is the quadratic compute cost and linear memory cost of attention, which becomes particularly relevant when scaling to long context lengths. Recently, several new architectures have been proposed that mitigate these limitations by using variants of linear attention \citep{katharopoulos2020transformers}. These state-space models (SSMs) such as Mamba \citep{gu2023mamba,dao2024mamba2}, RWKV \citep{peng2023rwkv,peng2024eagle}, and GLA \citep{yang2024gla} possess both highly parallelizable sequence mixing to ensure efficient training on GPUs while also possessing a recurrent formulation which enables $\mathcal{O}(1)$ memory and linear compute cost during autoregressive generation. Moreover, SSM models have been found to achieve scaling performance comparable to transformers in many domains. Notably, SSMs differ from conventional RNNs or their more advanced variants such as LSTMs \citep{schmidhuber1997lstm} and GRUs \citep{chung2014gru}, because, while they are recurrent, their recurrence can be parallelized and mapped efficiently to GPU hardware to make training and prompt infilling efficient~\citep{vaswani2017attention}. While pure SSM models appear to underperform on tasks involving in-context-learning \citep{grazzi2024mamba} and long-context retrieval \cite{park2024can}, when hybridized with attention blocks, they match or exceed the performance of pure transformer models while maintaining most of the inference efficiency of SSM models \citep{waleffe2024empirical,glorioso2024zamba,park2024mamba}.

Driven by the discovery of the scaling laws \citep{hestness2017deep,kaplan2020scaling,chinchilla}, model size has rapidly increased, driving concomitant explosions of both model training and inference costs. While continued scaling occurs at a rapid pace in order to reach the peaks of performance, smaller language models are also becoming rapidly more capable. Levels of performance that were previously thought to require models at the 100B-parameter scale or larger are now possible with models of with less than 10B parameters. For instance, leading 7B models \citep{llama3,jiang2023mistral,glorioso2024zamba} now outperform the original GPT-3 at many classic language modeling evaluations. These dramatic improvements in model quality have been driven primarily by a substantial increase in the quality and scale of the pretraining datasets used to train them. Moreover, many of these models are open-weights, enabling free study and customization, and have the additional advantage that they can be run locally on consumer GPUs or devices instead of on large GPU clusters in the cloud. However, despite the compelling advantages of SSM hybrids in terms of performance, until now the leading models have been pure transformer models. The reduced memory usage and reduced inference compute requirements of SSM-hybrid models is especially compelling for running models on-device, where both memory and compute are highly constrained. 

In this technical report, we release the Zamba2 series of models -- a 1.2B, 2.7B and 7.4B parameter suite of language models that achieve state-of-the-art performance across a wide range of language modeling evaluations in addition to leading inference and memory efficiency. The Zamba2-series models can achieve a peak speedup of 30-50\% time-to-first-token reduction as well as a 6x reduction in KV cache memory requirements over comparable transformer models due to their SSM-based architecture. We release the weights of all models under a highly permissive Apache 2.0 license. Moreover, unlike prior leading open-weights models, we also release our pretraining dataset for others to build upon.

\section{Architecture}

\begin{figure*}[t]
    \centering
    \subfigure[Zamba2-1.2B architecture]{
        \includegraphics[width=0.4\linewidth]{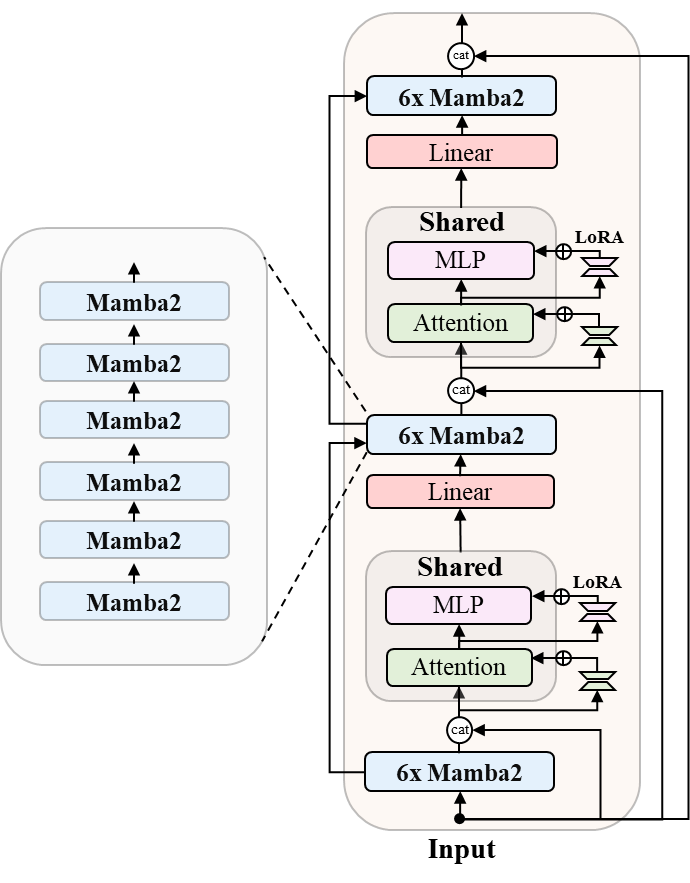}
    }
    \subfigure[Zamba2-2.7B and -7.4B architecture]{
        \includegraphics[width=0.4\linewidth]{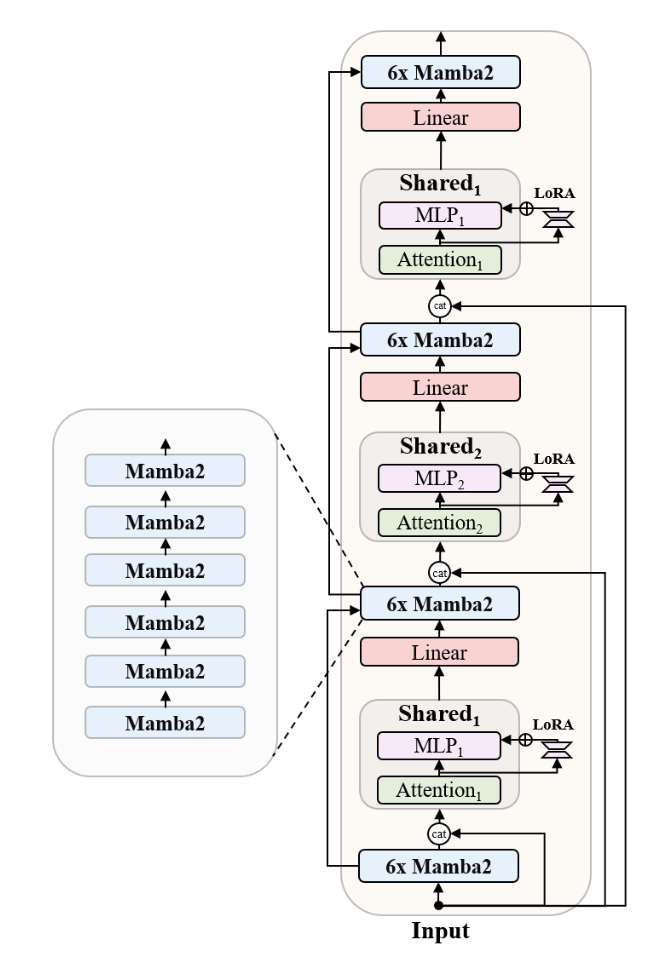}
    }
    \caption{Architecture diagrams for the 1.2B, 2.7B and 7.4B models. The 1.2B architecture differs in also including LoRAs on the shared attention blocks and only a single shared block. The single shared block for the 1.2B was used because the benefit of two alternating blocks was significantly less for the smaller model, since there are less total attention blocks. The 2.7B and 7.4B models lack the shared attention LoRAs because we only discovered that it was beneficial after training had commenced.}
    \label{fig:arch}
\end{figure*}

The Zamba2 architecture (Fig. \ref{fig:arch}) builds upon the innovations introduced in our previous Zamba1-7B model \citep{glorioso2024zamba}. There, we pioneered the use of a mamba backbone with shared attention blocks to optimize the performance per parameter. In our Zamba2 series, we performed a series of rigorous ablations to improve this architecture. Specifically, this led to the following improvements:
\begin{itemize}
    \item We switched from a Mamba1 \citep{gu2023mamba} to a Mamba2 \citep{dao2024mamba2} backbone. We found that Mamba2 has significantly higher throughput than an equivalently sized Mamba1, with approximately the same performance. This allows throughput to be traded for a larger state-size than is possible in Mamba1, which improves model performance overall.
    \item We use two alternating shared attention blocks instead of a single shared block as in Zamba1. This improves performance against parameter-matched baselines, with the additional benefit of using fewer FLOPs in inference, since there are fewer total attention layers, and training compared to the pure shared attention approach of Zamba1.
    \item We apply non-shared Low-Rank Adapters (LoRAs) \citep{hu2021lora} to the shared transformer blocks (Attention, MLP, or both) -- this offers each shared attention block additional expressivity, since the entire attention computation is not forced to be the same at different layers, at a small cost of additional parameters. 
    \item We apply Rotary Position Embeddings \citep{su2023rotary} to the shared attention blocks. While not strictly necessary for hybrid SSM-transformer models, we found that doing so improved performance, perhaps by providing an additional source of position information beyond simply the causal mask and the SSM cache. 
\end{itemize}

In our architecture ablations, we focused on maximizing the marginal loss decrease per parameter and secondarily finding the models with the minimum marginal FLOPs per parameter. We found that focusing on these two metrics as well as rigorously testing FLOP-matched and parameter-matched baselines significantly helped in performing architecture search. 

Due to various implementation and timing details, our Zamba2 series of models are slightly heterogeneous (Fig. \ref{fig:arch}). This is because we were exploring and running architecture search experiments in parallel with main model training. In particular, our 2.7B model lacks rotary position embedding in the attention, and our 1.2B model possesses LoRAs on both the shared attention and shared MLPs, and in addition only has one shared attention block, like the original Zamba1, rather than two alternating ones. This is because we found the benefit of two shared blocks to decrease at smaller scales, likely because of the fewer total attention layers.

\section{Pretraining}

The Zamba2 series of models were trained in two phases: a standard pretraining phase on primarily web data followed by an annealing phase consisting of a rapid decay of the learning rate over a mixture of web and higher-quality data. The 1.2B and 2.7B models were trained for 3T tokens. Due to compute and time limitations, we trained the Zamba2-7.4B model for 2T instead of 3T tokens. In the pretraining phase, for all models, we used the Zyda-2 dataset described in the following section. This dataset consists of heavily filtered web-text from a variety of sources, but especially including the FineWeb-Edu \citep{penedo2024fineweb} and DCLM \citep{li2024dclm} datasets, which are filtered by educational quality. We found that this provided significant boosts to the model's capabilities in factual knowledge recall and reasoning capabilities as exemplified by the MMLU and ARC evaluation metrics. 

Pretraining was conducted on our main compute cluster of 16 nodes, each consisting of 8xH100 SXM nodes with 3.2Tbps Infiniband. To train the 2.7B and 1.2B models, data parallelism was all that was required. To train the 7.4B model we utilized two-way tensor parallelism which we implemented for both Mamba2 and our shared attention block. We used the ZeRO-1 \citep{rajbhandari2020zero} distributed optimizer to shard optimizer states, and activation checkpointing. We trained with a base sequence length of 4096 tokens. All models of the Zamba2 series use the Mistral-7B tokenizer. The 1.2B model took approximately 20 days to train. The 2.7B model took approximately 40 days to train and the 7.4B model took approximately 60 days to train. By careful hyperparameter selection, architecture optimization, gradient clipping, and weight decay, we were able to avoid any instabilities during optimization. We used the Adam optimizer \citep{kingma2014adam} for all pretraining.

Following phase-1 pretraining, we performed annealing on all of the models of the Zamba2 series. Our annealing phase broadly follows that implemented in Zamba1 and is inspired by miniCPM \citep{hu2024minicpm}. We performed a rapid learning rate decay over an annealing dataset consisting of high-quality factual, math, code, and instruction-following datasets. This dataset was mixed with our phase-1 pretraining dataset at a replay ratio of 60\% phase-1 data and 40\% annealing data. Our annealing dataset consisted of approximately 50B tokens, and we annealed for two epochs resulting in a total 100B tokens for the annealing phase. 

For phase-1 pretraining, we performed a cosine decay schedule starting from an initial maximum learning rate. For the annealing phase, we re-warmed the learning rate back up to approximately mid-way between the initial and final learning rates of the phase-1 pretraining before quickly decaying to a very low learning rate. During annealing, we maintain a replay fraction of $60\%$ from the phase-1 dataset to mitigate catastrophic forgetting. In both cases we used cosine learning rate schedules. We found that the precise schedule used makes little difference to the outcome, while the specific values of the initial and final learning rates for both phase-1 pretraining and annealing are critical. For more practical details on annealing schedules and replay fractions, we refer to \cite{anthony2024zcookbook} and \cite{ibrahim2024simple}.

\section{Datasets}

\begin{figure}
    \centering
    \includegraphics[width=\linewidth]{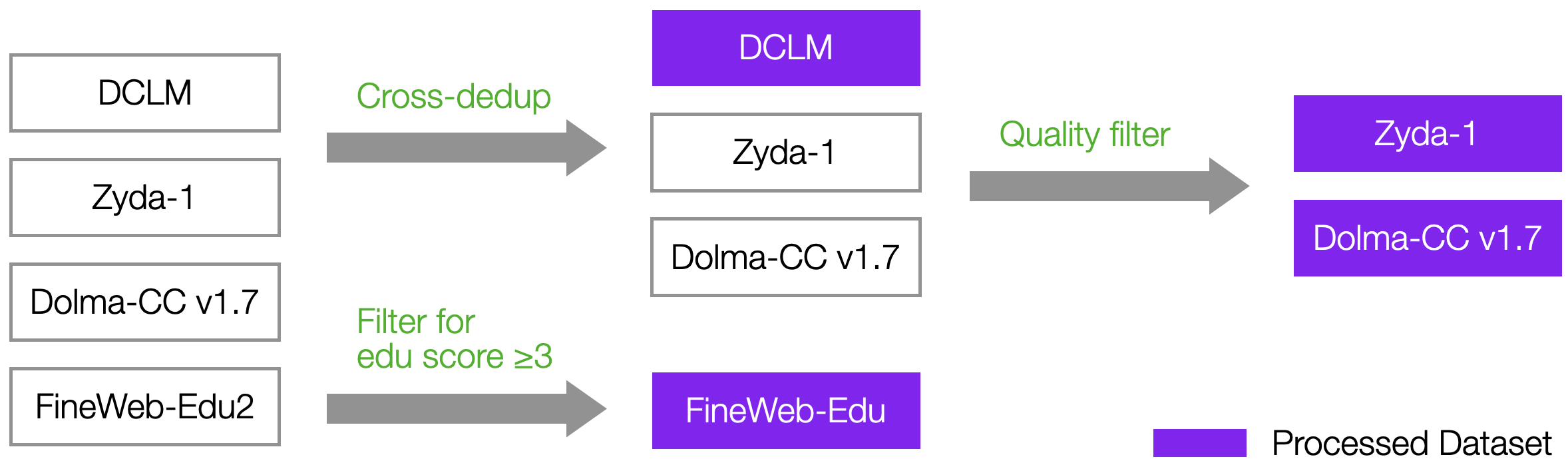}
    \caption{Pipeline for producing the Zyda-2 dataset. Zyda-2 comprises four component datasets: Zyda-1, DCLM, FineWeb, and Dolma. We cross-deduplicated all dataests against each other. For Zyda-1 and Dolma we also performed model-based quality filtering using Nvidia's Nemo.}
    \label{Zyda2_creation}
\end{figure}

For our pretraining dataset, we used Zyda-2 \citep{zyda2}, a dataset of 5 trillion tokens consisting of FineWeb-Edu3 \citep{penedo2024fineweb}, DCLM \citep{li2024dclm}, Zyda-1 \citep{tokpanov2024zyda}, and Dolma-common-crawl \citep{soldaini2024dolma}. We performed model quality-based filtering on the Dolma and Zyda-1 subsets and full cross-deduplication across all component datasets (Fig. \ref{Zyda2_creation}). 

\begin{figure}
    \centering
    \includegraphics[width=\linewidth]{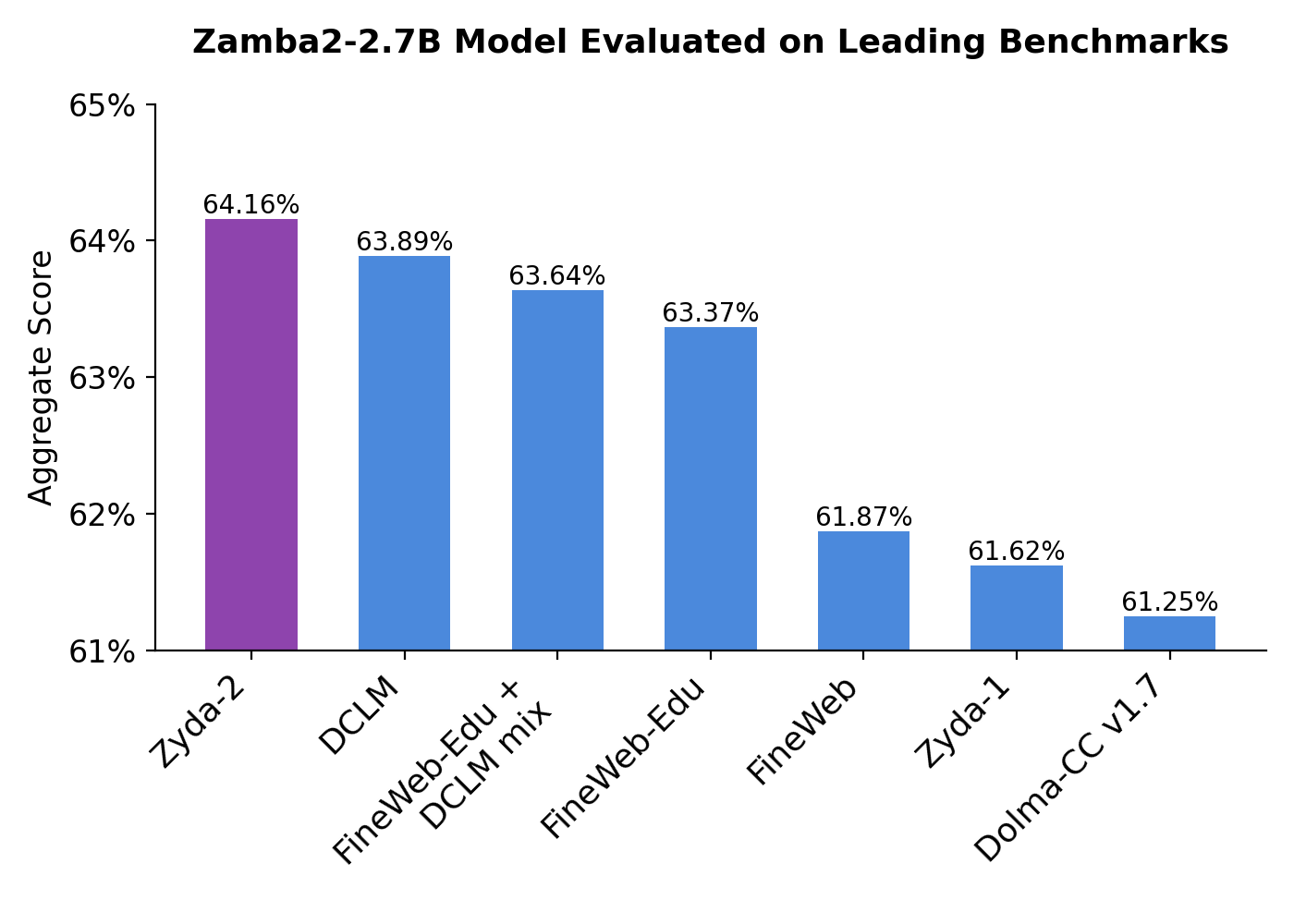}
    \caption{The performance of Zyda-2 vs other leading language modelling datasets. Reported is the average score on a set of standard language modelling evaluation tasks for annealing Zamba2-2.7b on each dataset. We followed \cite{blakeney2024does}'s annealing ablation protocol to measure performance instead of training models from scratch because we observed significantly higher signal with this approach.}
    \label{zyda2_performance}
\end{figure}

We found that our Zyda-2 dataset outperformed previous state-of-the-art datasets in annealing ablation tests (Fig. \ref{zyda2_performance}). This is because Zyda-2 uses additionally filtered versions of these datasets as well as other high quality sets such as Zyda-1, which means that Zyda-2 benefits from an ensembling effect of putting together multiple sources which can help compensate for the weaknesses of each specific dataset.

While we trained on pure Zyda-2 for the 1.2B and 2.7B models, for the 7.4B model we augmented the dataset with 10\% StarCoder \citep{li2023starcoder} in order to improve the model's coding capabilities, since we believed such capabilities are more important for generalist models of the 7B scale rather than smaller models.

\section{Model Performance}

\subsection{Quality}

\begin{table*}[ht]
  \centering
  \caption{Performance of Zamba2 series of models vs competitors. The number of evaluation shots is in parentheses, and each maximum score at a given model scale is bolded.}
  \label{table1}
  \begin{tabular}{lcccccccc}
    \toprule
    Model & MMLU (5) & Arc-Easy (0) & Arc-Challenge (25) & Hellaswag (10) & Piqa (0) & Winogrande (0) & Boolq (0) & OBQA (0) \\
    \midrule
    \multicolumn{9}{c}{\textbf{1-2B Scale}} \\
    Zamba2-1.2B & \textbf{43.1} & 72.7 & 45.9 & 70.9 & \textbf{78.3} & \textbf{68} & \textbf{75.0} & \textbf{43.6} \\
    Gemma2-2B & 32.8 & 72.4 & 41.6 & \textbf{71.4} & 78.3 & 65.1 & 69.6 & 39.8 \\
    Llama3.2-1.2B & 36.83 & 60.31 & 36.18 & 63.68 & 74.48 & 60.3 & 63.67 & 37.2 \\
    StableLM-1.6B & 30.8 & 68.3 & 39.8 & 68.9 & 76.5 & 63.7 & 65.6 & 38.8 \\
    SmolLM-1.7B & 27.65 & \textbf{73.5} & \textbf{46.2} & 65.7 & 76.06 & 60.93 & 65.3 & 42.0 \\
    \midrule
    \multicolumn{9}{c}{\textbf{2-3B Scale}} \\
    Zamba2-2.7B & \textbf{55.97} & \textbf{80.13} & \textbf{60.0} & \textbf{76.35} & \textbf{80.36} & \textbf{73.24} & 74.25 & \textbf{46.4} \\
    Llama3.2-3B & 54.01 & 71.6 & 52.6 & 73.6 & 76.0 & 69.9 & 72.8 & 43.2 \\
    Gemma2-2.6B & 51.3 & 80.1 & 54.4 & 73.0 & 77.8 & 70.9 & 72.5 & 40.0 \\
    StableLM-3B & 40.3 & 68.3 & 46.33 & 68.9 & 76.5 & 63.7 & \textbf{74.4} & 38.8 \\
    
    \midrule
    \multicolumn{9}{c}{\textbf{7-8B Scale}} \\
    Zamba2-7B & \textbf{67.2} & \textbf{82.0} & \textbf{68.9} & \textbf{81.5} & 81.0 & \textbf{77.3} & \textbf{84.1} & 48.2 \\
    Llama3.2-8B & 65.18 & 77.61 & 57.85 & 79.13 & 80.96 & 73.24 & 81.16 & 45.0 \\
    Mistral-7B & 62.2 & 79.59 & 61.43 & 81.07 & \textbf{82.26} & 73.88 & 83.64 & 44.2 \\
    Gemma-7B & 62.9 & 80.7 & 61.1 & 80.46 & 81.1 & 73.8 & 83.12 & 45.2 \\
    FalconMamba-7B & 59.86 & 81.2 & 58.4 & 80.2 & 80.2 & 75.4 & 83.2 & \textbf{48.6} \\
    \bottomrule
  \end{tabular}
\end{table*}

Models of the Zamba2 series achieve leading performance compared against models of their weight class as measured by standard language model evaluation metrics (see Table \ref{table1}). We believe that this performance is attributable to the quality of our pretraining and annealing datasets, as well as the improved architecture of Zamba2, in particular when compared to pure transformer baselines. The Zamba2 series showcases the fact that highly-performant small models can be made with significantly smaller compute and token budgets than were previously thought necessary for a given level of performance.

\begin{figure}
    \centering
    \includegraphics[width=\linewidth]{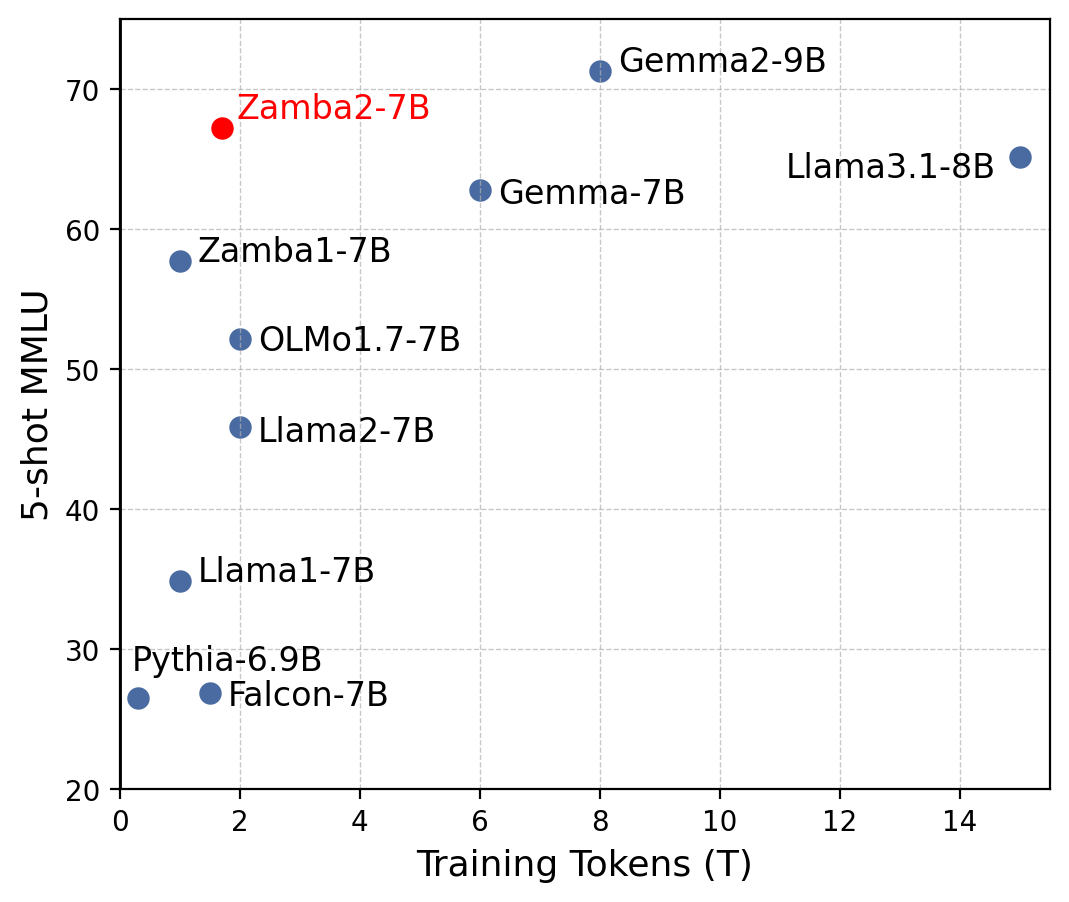}
    \caption{Performance (in 5-shot MMLU) vs the number of tokens used for training. We observe a fairly clear sigmoidal curve of performance vs training tokens for current leading transformer models with Zamba2 and Gemma2 being clear outliers. We believe Gemma2 is an outlier because of its use of distillation from a larger model, while Zamba2 outperforms due to its architecture.}
    \label{fig:mmlu}
\end{figure}

Perhaps the best way to demonstrate the strong evaluation performance of the Zamba2 architecture is to measure the performance per training token, which is a measure both of dataset and of architecture quality. As shown in Fig. \ref{fig:mmlu}, the Zamba2 series performs extremely strongly on this metric compared to the most competitive models of their scale. Given that our dataset is open (Zyda-2) while the datasets of almost all comparable models are closed, we believe it is unlikely that our dataset alone gives us a strong advantage and thus conclude that Zamba2's superior evaluation performance is largely due to the Zamba2 architecture.

\subsection{Inference performance}

Due to the improved throughput of the Mamba2 block over standard transformer blocks, Zamba2 models achieve significantly higher throughput and lower latency (time-to-first-token and time-per-output-token) compared to existing transformer models at comparable scales. Mamba2 blocks have approximately 4$\times$ the throughput of standard transformer blocks, and in-line with our philosophy of maximizing performance per parameter, we exploit these FLOP savings to add our shared attention layers, boosting model quality while still remaining significantly faster and more performant than comparable transformer models.

\begin{figure*}[t]
    \includegraphics[width=\linewidth]{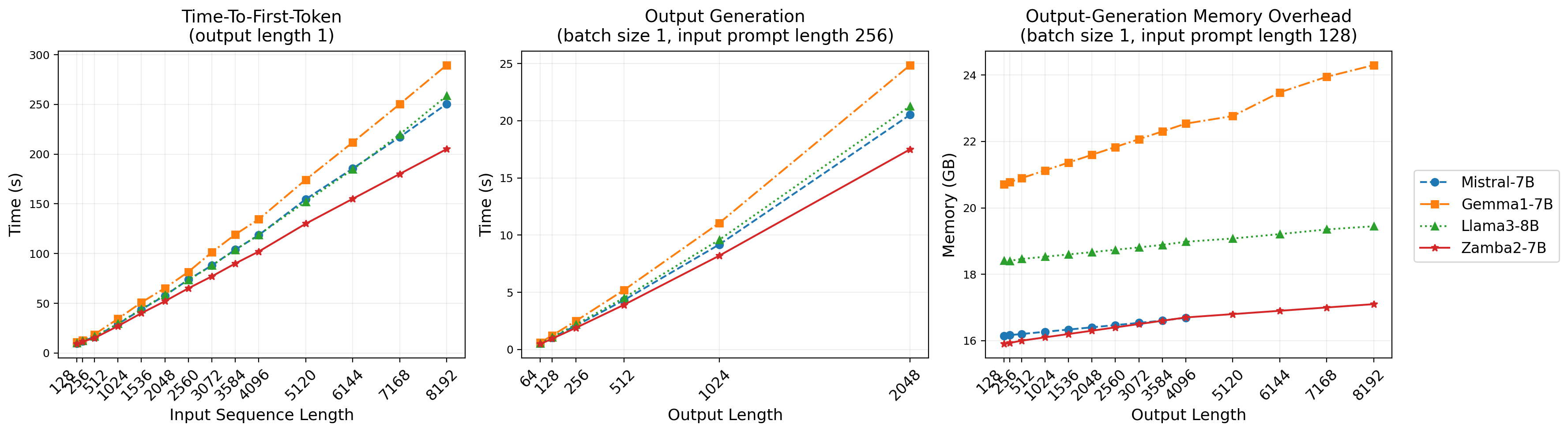}
    \caption{Performance characteristics of the Zamba2-7B model vs competing models: time-to-first-token \textit{(left)}, generation throughput \textit{(middle)} and memory utilization \textit{(right)}.}
    \label{fig:perf1}
    \vspace{-0.5cm}
\end{figure*}

\begin{figure*}[t]
    \includegraphics[width=\linewidth]{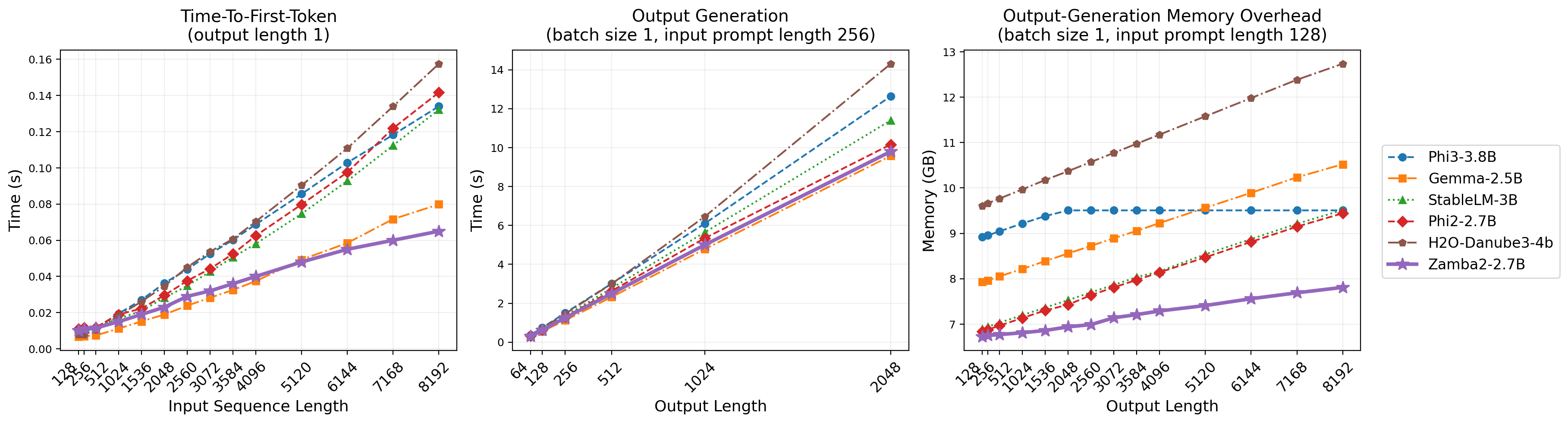}
    \caption{Performance characteristics of the Zamba2-2.7B model vs competing models: time-to-first-token \textit{(left)}, generation throughput \textit{(middle)} and memory utilization \textit{(right)}.}
    \label{fig:perf2}
\end{figure*}

\begin{figure*}[t]
    \includegraphics[width=\linewidth]{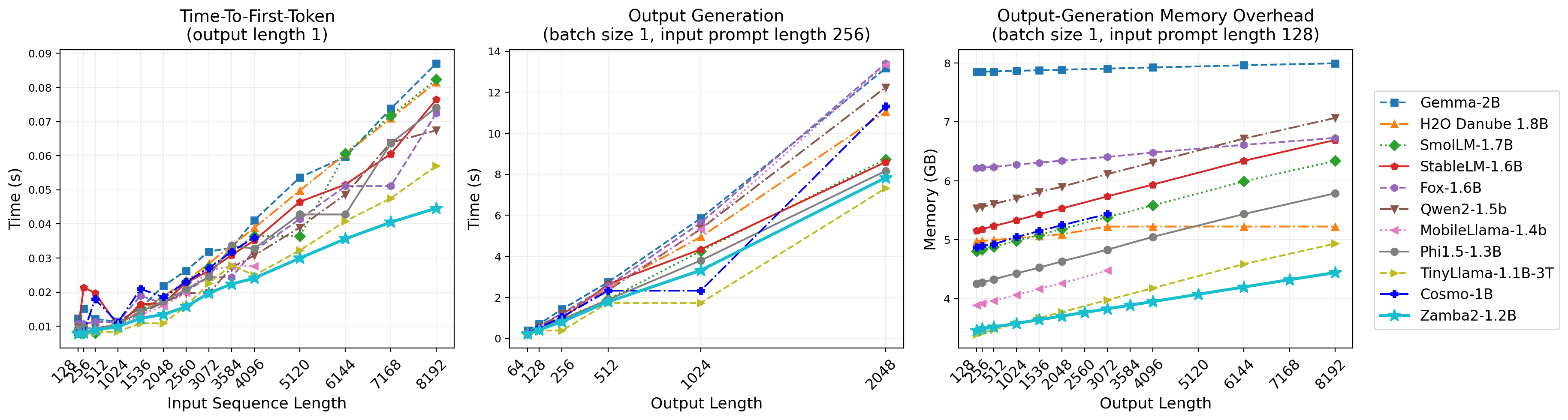}
    \caption{Performance characteristics of the Zamba2-1.2B model vs competing models: time-to-first-token \textit{(left)}, generation throughput \textit{(middle)} and memory utilization \textit{(right)}.}
    \label{fig:perf3}
\end{figure*}

Additionally, since the SSM backbone has a fixed-size state cache, at long context lengths our Zamba2 series of models require significantly less memory for generation than transformers. Zamba2 models must only store KV caches for each invocation of the shared attention layers in our hybrid architecture which, at a 1:6 ratio of Mamba2 to attention, reduces our KV cache requirements by a corresponding 6$\times$ compared to pure transformers. Such a reduction in memory usage is significant for both serving long contexts in the cloud and on-device applications where device memory may be very limited. See Figures~\ref{fig:perf1}-\ref{fig:perf3} for inference performance benchmarks.

Finally, we performed a number of optimizations of the core operations to boost training and inference performance. These include correctly sizing model dimensions to be amenable to GPU hardware \citep{anthony2024codesign} as well as performing extensive kernel fusion and general removal of unnecessary operations in the core model code.

\section{Post-Training}

\subsubsection{Instruct Tuning}

In addition to releasing annealed checkpoints, we also release instruction-tuned versions of our base Zamba2 models which are specialized for instruction-following and AI-assistant chat use cases. We have found performance of our instruct models to be strongly competitive with the existing official instruct finetunes of competitor models (for each of the Zamba2 series), using only open-source finetuning datasets and methodologies (see Table \ref{table_postt}).

Our instruction tuning approach consists of a phase of supervised finetuning on instruct and chat data, followed by multiple iterations of DPO \citep{rafailov2024dpo}. The supervised finetuning dataset consisted of about 4M samples, including OpenHermes \citep{OpenHermes}, UltraChat \citep{ding2023ultrachat}, Infinity-Instruct \citep{InfinityInstruct} and Llama-3-Magpie-Pro \citep{xu2024magpie}. For the secondary DPO phase, we used over 200k samples mostly from UltraFeedback \citep{cui2024ultrafeedback}, Orca-DPO pairs \citep{orca_dpo_pairs} (sourced from OpenOrca \citep{OpenOrca}), and OpenHermes-Preference \citep{open_hermes_preferences} datasets. We finetuned our models to follow the standard ChatML template. We found that instruction-tuning significantly improved the usability of our models as measured by metrics such as IFeval \citep{zhou2023ifeval} and MT-bench \citep{zheng2023mtbench}. We found that our models are relatively robust to changes in format compared to alternative models which tend to suffer drastic drops in evaluation scores if the chat template is different than expected.

\begin{table}[ht]
  \centering
  \caption{Performance of Zamba2 Instruct and other models at similar scales.}
  \label{table_postt}
  \begin{tabularx}{\linewidth}{Xcc}
    \toprule
    Model & MT Bench & IFEval \\
    \midrule
    \multicolumn{3}{c}{\textbf{1.2B Scale}} \\
    Zamba2-1.2B-Instruct & 5.45 & 41.3\\
    SmolLM-1.7B-Instruct & 3.89 & 53.6 \\
    Llama3.2-1.2B-Instruct & 5.41 & 57.4 \\
    StableLM-1.6B-Chat & 4.66 & 31.4 \\
    H2O-Danube2-1.8B-Chat & 4.39 & 25.6 \\
    \midrule
    \multicolumn{3}{c}{\textbf{2.7B Scale}} \\
    Zamba2-2.7B-Instruct & 6.95 & 48.0 \\
    Llama3.2-3B-Instruct & 7.37 & 75.3 \\
    Gemma2-2.6B-Instruct & 4.97 & 55.8 \\
    StableLM-Zephyr-3B & 5.52 & 37.2 \\
    
    \midrule
    \multicolumn{3}{c}{\textbf{7.4B Scale}} \\
    Zamba2-7B-Instruct & 7.42 & 69.94 \\
    Llama3.1-8B-Instruct & 8.19 & 78.00 \\
    Mistral-7B-Instruct-v0.3 & 7.37 & 54.2 \\
    Gemma-7B-it & 2.99 & 37.3 \\
    \bottomrule
  \end{tabularx}
\end{table}

\begin{figure}
\centering
\includegraphics[width=0.95\linewidth]{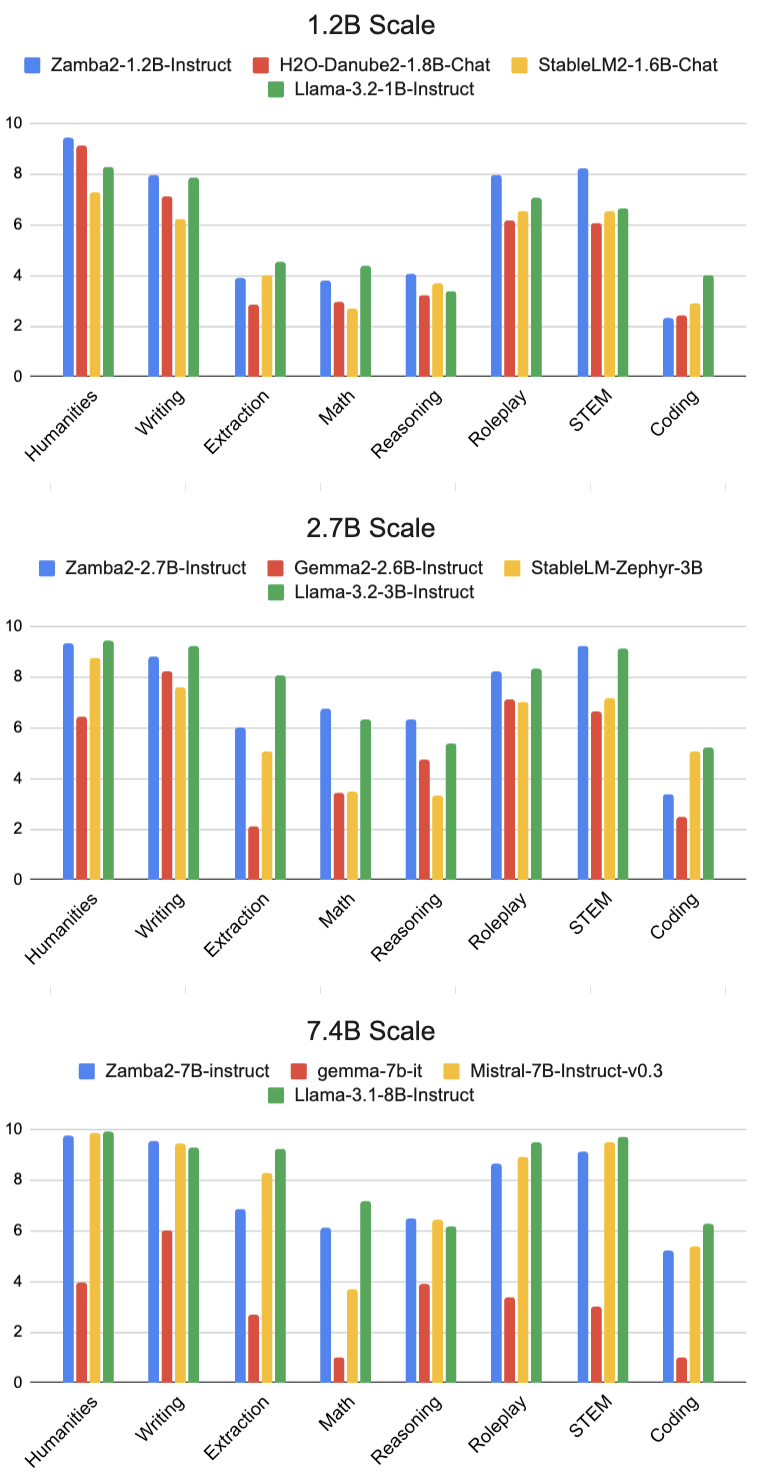}
\caption{MT Bench scores per subject of Zamba2 Instruct and other models across various scales.}
\end{figure}

\subsubsection{Context Extension}
Zamba2 was trained on a fixed context window of 4096 tokens, but it shows a remarkable degree of generalization to longer contexts when measured in terms of LM loss. 
However, when evaluated on tasks such as passkey retrieval (also known as ``retrieval of a needle from a haystack''), the model fails shortly outside its context window. This can further be remedied by modifying the rotary position embedding inside the shared attention blocks.
Specifically, we use the ``NTK-aware'' scaling introduced by \cite{blocntkaware}. The idea is to rescale the angle variable in the rotary embedding $\theta_d$ as 
\begin{equation}
\theta'_d = \frac{\theta_d}{s^{d_\text{emb}/(d_\text{emb}-1)}},
\end{equation}
where $d_\text{emb}$ denotes the embedding dimension and $s$ is a scaling factor we chose to be 16. In conventional transformers, this factor is typically taken to be the ratio of the target maximum context length to the current maximum context length, but in our case this identification does not strictly hold. 
With this modification of the rotary embedding, the 7B model's effective context window can be extended up to 17000 tokens without additional training. The performance we see is consistent with observations in other studies \citep{Yarn}, where they noticed that NTK-aware scaling is very well suited for context extension without additional finetuning. 

\begin{figure}
\centering
\includegraphics[width=0.9\linewidth]{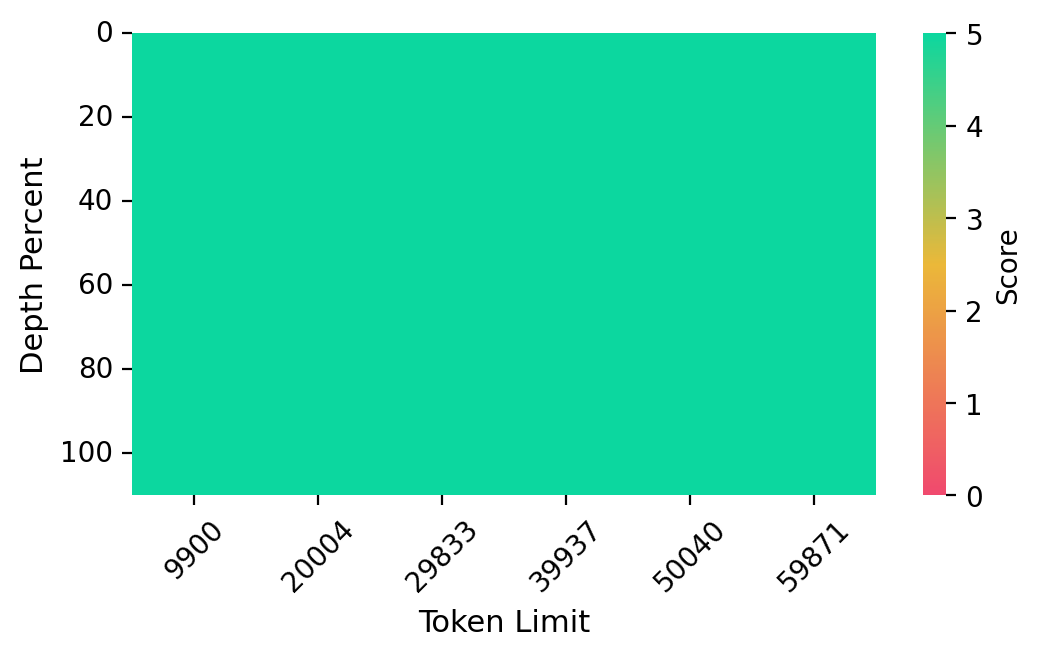}
\caption{Zamba2-2.7B's performance on passkey retrieval. ``Token Limit'' denotes the length of the sequence within which the needle query is embedded. The needle query can be embedded at all locations within the long context, and this depth of embedding is denoted ``Depth Percent''.}
\label{fig:ctx_extend}
\end{figure}

In Zamba2-2.7B, where no positional embedding was used, we observed that simple finetuning on long-context pretraining data of length up to 65536 extends the model's ability to perform accurate passkey retrieval up to this extended context window (Fig. \ref{fig:ctx_extend}). The sequences in the data that was used for this extension were simply concatenated sequences in the pretraining data up to the target context length. We found that it helped to perform this finetuning phase with a step-wise curriculum, where at every 100 steps the context length was doubled starting from 4096 until we reached 65536. We found that the primary bottleneck on further context-length extensions for this model was the high RAM cost of training at extremely long sequence lengths.

\subsection{Quantization}

We have experimented with finetuning Zamba2 models for specific tasks, e.g. text summarization, using QLoRA~\citep{dettmers2023qlora} -- a method for efficient finetuning that quantizes the weights of the base model, freezes them and finetunes only the weights of added LoRAs \citep{hu2021lora}. 
We found that, largely, standard quantization approaches generalize well to hybrid attention-SSM models from transformers without requiring significant modification. The primary special considerations when quantizing SSMs is the importance of keeping the SSM and convolutional state and sensitive SSM matrices such as the $A$ matrix and $dt$ projection in higher precision. The remainder of the parameters in the SSM block can be successfully quantized.

We quantized the linear layers of the Zamba2-2.7B instruct model (excluding the embedding and unembedding layers) to 4-bit precision, added LoRAs to the linear layers in the attention blocks of Zamba2, and performed supervised finetuning of the LoRAs on a small custom dataset. We found that a dataset of several hundred thousands high-quality samples is sufficient to achieve satisfactory performance. Subsequently we quantized the LoRA parameters to 4-bit as well. 
Quantizing the Zamba2-2.7B to 4-bit precision reduces the memory footprint from 5.38 GB to 1.55 GB; with 4-bit quantized LoRA parameters the final model is at just 1.7 GB. This combination of high performance and small size enables the Zamba2 series of models to be deployed effectively in a wide variety of on-device environments. Training a set of LoRAs offers an efficient and lightweight solution for specializing the one base model for several tasks. We used the HuggingFace libraries BitsAndBytes \citep{dettmers2022bitsandbytes} and PEFT \citep{peft} for quantization and finetuning.

\section{Related Work}

The space of small language models has become increasingly crowded as the general potential of such models becomes widely recognized. A number of open-weight transformer model families achieve strong performance at small scales, such as the Llama3 series \citep{dubey2024llama3herdmodels}, Mistral models \citep{jiang2023mistral}, and the Gemma1 and Gemma2 series of models \citep{team2024gemma}; these models have been our quality benchmarks to compare to and improve upon while developing our Zamba2 series of models.

Additionally, a number of groups have attempted to scale up state-space model architectures which has resulted in a growing number of competitive models to transformers. For instance, \citet{waleffe2024empirical} pretrain a suite of pure Mamba and Mamba2 hybrid models and report that the hybrid models can outperform transformers when trained on sufficiently many tokens up to a 7B model scale. Recently, a pure Mamba 7B model with high quality \citep{zuo2024falconmamba} was released which demonstrates that at sufficient pretraining token count even a pure SSM catches up and can seemingly outperform transformers, although the asymptotic quality of a pure SSM vs hybrid was not rigorously tested. Additionally, AI21 released a series of scaled up Mamba/MoE hybrid models \citep{lieber2024jamba, jambateam2024jamba1p5} with strong performance at their scales, following on from our smaller scale work \citep{anthony2024blackmamba}. Additionally, other SSM-style architectures such as RWKV \citep{peng2023rwkv,peng2024eagle} have been showing successful scaling capabilities up to the 7B range.

While there remain few truly rigorous studies of SSMs and SSM hybrids vs transformers at scale, it is broadly clear that SSMs can scale and deliver performance comparable to extremely strong transformer baselines if trained with sufficiently many resources and high quality datasets. This discovery opens the door for further innovations in architecture which challenge the existing transformer-only status-quo and potentially leads to significantly higher performing architectures arising in the near future.

\section{Discussion}

The performance of language models at small parameter sizes has risen dramatically over the past few years, with capabilities that were previously thought to require hundreds of billions of parameters being achieved with less than 10 billion. This dramatic increase in capability has seemingly been driven by a few relatively simple factors -- namely a vast increase in the quality and scale of pretraining datasets. From a few hundred billion tokens of completely unfiltered web-crawl, to the order of ten-trillion token datasets comprised of extensively filtered and vetted tokens carefully designed to maximize specific desirable capabilities of the model. The ``Chinchilla scaling laws'' \citep{chinchilla} showed how in many cases data quantity was significantly more important than previously recognized, and that existing models of the time were dramatically undertrained. Moreover, by accounting for inference efficiency, models are now trained for dramatically longer on more tokens than the Chinchilla scaling laws would predict, because performance continues to improve monotonically, although strongly sublinearly, in the training dataset size. Small models today are trained for trillions of tokens on high-quality datasets, some \citep{abdin2024phi} including large amounts of synthetic data. 

In our previous paper, we speculated upon the necessary ingredients to reach the frontier, and our current work confirms that simply training on twice the total tokens and on a significantly higher-quality pretraining dataset is all that is required -- for now. The interesting question then becomes what the methods that will drive the next frontier of performance. 

One promising direction are improvements to the model architecture. In many experiments we have observed that in rigorously parameter and FLOP-matched ablations, the Zamba2 architecture significantly outperforms standard Llama-style transformer baselines. We speculate that given the strongly sublinear returns to training on additional tokens, in the current regime of significantly-Chinchilla-overtrained models, the constant improvement factor given by a novel architectural innovations can outweigh extremely large FLOP differentials in pretraining. We see this empirically when comparing the FLOP-vs-performance graphs of Zamba2 vs other transformer models. We expect that the Zamba2 architecture is still far from the optimum and that there is significant room for optimization of the marginal loss per parameter.

In addition, we believe there are still improvements which can be made in pretraining and annealing dataset design. The Phi series of models \citep{abdin2024phi,li2023textbooks} have highlighted one possible path of including industrial quantities of synthetic data directly into the pretraining mix. This approach holds promise to augment and specialize model capabilities towards those we tend to care about, as well as to address weaknesses in existing models. Beyond this, some recent models, such as Gemma2-9B have described using distillation from existing larger models to boost performance \citep{gemmateam2024gemma2}. While more expensive in FLOPs than standard pretraining, this could provide another strong avenue of performance improvement similar to the effects of Chinchilla-overtraining. Overall, it is highly likely that the performance of small, open-weight models can be strongly improved, and this will have the effect of democratizing powerful LLMs and making them accessible and abundant throughout the economy. 

\clearpage

\section*{Author Contributions}
\textbf{Paolo} — Lead post-training. Contributed to core infrastructure. Lead Huggingface conversion and release.  Contributed to pretraining.

\textbf{Quentin} — Lead model optimization and inference. Contributed to core infrastructure. Lead cluster management and maintenance. Contributed to evaluations. Contributed to pretraining.

\textbf{Yury} — Lead pretraining dataset (Zyda-2) creation. Contributed to core infrastructure and cluster maintenance. 

\textbf{Anna} — Lead quantization. Contributed to post-training. 

\textbf{Vasu} — Lead context length extension.

\textbf{James} — Contributed to architecture search experiments.

\textbf{Jonathan} — Contributed to architecture search experiments.

\textbf{Beren} — Overall project lead. Lead annealing dataset creation and annealing phase. Lead evaluations. Contributed to core infrastructure. Contributed to architecture search experiments. Contributed to post-training and pretraining datasets. Contributed to cluster management. Contributed to pretraining.

\bibliographystyle{tmlr}
\bibliography{main}


\end{document}